\documentclass[11pt]{preprint}

\usepackage[font=normalsize,labelfont=bf,justification=justified]{caption}
\usepackage{xspace}
\usepackage[super,comma,sort&compress]{natbib}
\usepackage{booktabs}
\usepackage[colorlinks=true]{hyperref}
\usepackage{graphicx}
\usepackage{multicol}

\newcommand{\DermpathNet}{\emph{DermpathNet}\xspace}

\title{Establishing dermatopathology encyclopedia \DermpathNet with Artificial Intelligence-Based Workflow\footnotemark[3]}
\date{}

\author[1,$\dag$]{Ziyang Xu}
\author[2,$\dag$]{Mingquan Lin}
\author[2]{Yiliang Zhou}
\author[2]{Zihan Xu}
\author[3]{Seth J. Orlow}
\author[3,4]{Shane A. Meehan}
\author[1,4,*]{Alexandra Flamm}
\author[1,4,*]{Ata S. Moshiri}
\author[2,*]{Yifan Peng}
\affil[1]{Perelman Department of Dermatology, NYU Grossman School of Medicine, New York, USA}
\affil[2]{Department of Population Health Sciences, Weill Cornell Medicine, New York, USA}
\affil[3]{Division of Dermatopathology, Mount Sinai Health, New York, USA}
\affil[4]{Division of Dermatopathology, Perelman Department of Dermatology, NYU Grossman School of Medicine, New York, USA}

\affil[*]{Corresponding author(s): \url{alexandra.flamm@nyulangone.org}, \url{ata.moshiri@nyulangone.org}, \url{yip4002@med.cornell.edu}}

\affil[$\dag$]{These authors contributed equally to this work}

\begin{document}

\maketitle
\footnotetext[3]{Accepted by \textit{Scientific Data}.}

\begin{abstract}
Accessing high-quality, open-access dermatopathology image datasets for learning and cross-referencing is a common challenge for clinicians and dermatopathology trainees. To establish a comprehensive open-access dermatopathology dataset for educational, cross-referencing, and machine-learning purposes, we employed a hybrid workflow to curate and categorize images from the PubMed Central (PMC) repository. We used specific keywords to extract relevant images, and classified them using a novel hybrid method that combined deep learning-based image modality classification with figure caption analyses. Validation on 651 manually annotated images demonstrated the robustness of our workflow, with an F-score of 89.6\% for the deep learning approach, 61.0\% for the keyword-based retrieval method, and 90.4\% for the hybrid approach. We retrieved over 7,772 images across 166 diagnoses and released this fully annotated dataset, reviewed by board-certified dermatopathologists. Using our dataset as a challenging task, we found the current image analysis algorithm from OpenAI inadequate for analyzing dermatopathology images. In conclusion, we have developed a large, peer-reviewed, open-access dermatopathology image dataset, \DermpathNet, which features a semi-automated curation workflow.
\end{abstract}

\section*{Background \& Summary}

Clinicians, as well as dermatology and pathology trainees, often struggle to access high-quality, open-access dermatopathology image datasets for diagnostic or educational purposes~\cite{Shahriari2017-pg, Wen2022-oz, Lee2015-kq, Rinck2023-mb, Feit2005-pg, Ward2024-le}. Physical or digital slides used for teaching are typically proprietary to a specific institution and may include patient identifiers, restricting their broader use\cite{Brown2015-zd, Holub2023-hp, Moore2015-gl}. Some sites, such as VisualDx, offer a limited collection of dermatopathology images but require a user subscription\cite{Vardell2012-vl}. Others, like Pathologyoutline.com, provide only a small number of images illustrating classic disease features\cite{Pernick2021-wb}. While PathPresenter offers high-resolution digital collections of scanned slides that can be uploaded by any user, not all cases on the platform are peer-reviewed, which may affect reliability\cite{Schuffler2021-ik}.

In this study, we developed a pipeline to accelerate the automated construction of a dermatopathology image dataset using the PubMed Central Open Access Subset (PMC-OA) as our source\cite{comeau2019pmc}. PMC-OA is a free, peer-reviewed, open-access platform that hosts an extensive range of biomedical information\cite{Ossom_Williamson2019-hy}. Specifically, we curated and cataloged images from PMC-OA that correspond to common benign and malignant cutaneous neoplasms, aiming to create a comprehensive dermatopathology dataset for educational purposes, cross-referencing, and machine learning applications. Our approach leveraged a deep learning (DL)-enabled algorithm to facilitate the identification and organization of relevant images. Additionally, we evaluated and compared the precision and recall metrics of traditional keyword-based retrieval against those of our DL-based retrieval algorithms.


\section*{Methods}\label{methods}

\subsection*{Constructing DermPath Lexicon}

We began by compiling a comprehensive list of benign and malignant cutaneous neoplasms to serve as our diagnoses of interest for this proof-of-concept study. We first compiled an initial lexicon with 12 categories, based on input from subject matter experts (SMEs). Next, we searched for symptoms in the UMLS Metathesaurus (version 2023AB) in the English language, using vocabulary sources of ``SNOMEDCT\_US'' and ``MeSH''. We included concepts that can be strictly matched in UMLS, along with their narrower concepts (children).

\subsection*{Collecting candidate images from PMC-OA}

Using the PubMed API (Entrez Programming Utilities\cite{sayers2022general}), we then searched for PMC-OA articles containing relevant keywords in their titles or abstracts. For instance, the query term for identifying articles related to Angiokeratoma was ``Angiokeratoma {[}Title/Abstract{]}.'' The E-utilities tool retrieved the PubMed Central ID (PMCID) for each relevant article, which serves as the unique reference number within the PMC-OA repository.

Next, we used these PMCIDs to obtain the full-text articles in XML format, which adheres to the NISO standard DTD for article publishing\cite{Beck2011-pz}. We then parsed the XML files to extract figures and their corresponding legends. When figures were presented as compounds containing multiple panels, we separated them into individual panels using an existing deep-learning model developed by Yao et al\cite{Yao2021-ik}.

\subsection*{Determine dermatopathology-specific images using a hybrid approach}

\begin{figure}
\centering
\includegraphics[width=.4\textwidth]{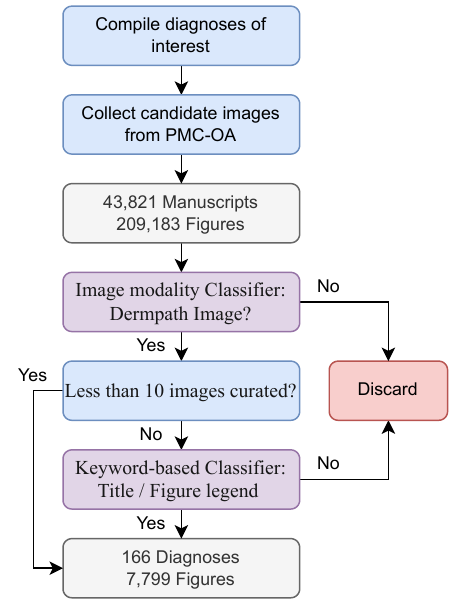}
\caption{Overview of the Artificial Intelligence empowered workflow for semi-automated retrieval of dermatopathology images from PubMed Central (PMC).\label{fig:overview}}
\end{figure}

Because the majority of figures retrieved from PMC-OA articles were not dermatopathology images, we developed a hybrid algorithm to accurately identify the relevant image modality. This approach combines a deep learning-based image modality classifier with keyword-based retrieval from the figure captions (\textbf{Figure~\ref{fig:overview}}).

\paragraph{Development of image modality classification (DL classifier).} 

First, we trained a deep learning-based image modality classifier using the ImageCLEF 2016 dataset (\textbf{Table \ref{tab:clef}})\cite{deHerrera2016overview}. Our model architecture was based on DenseNet-121, pre-trained on ImageNet\cite{deng2009imagenet}. We replaced the original classification layer with a fully connected layer followed by a sigmoid activation function, enabling the model to output the probability of an input image being a pathology image. All images were resized to 224 $\times$ 224 $\times$ 3 to standardize input dimensions.

We grouped the ImageCLEF 2016 image modality labels into a binary classification scheme for training. Specifically, the “Dermatopathology” class (referred to as ‘light microscopy’ in \url{https://www.imageclef.org/2016/medical}) was designated as the positive category, and all other modality classes (including radiology, microscopy, and generic biomedical illustrations) were combined into a single “non-pathology” category. This enabled us to train the DL classifier as a binary model.

The models were implemented in Keras with a TensorFlow backend and optimized using the Adam optimizer with a learning rate of $5 \times 10^{-5}$. Stochastic image augmentation was applied at three levels: (1) random rotation between 0\textdegree and 10\textdegree, (2) random translation along the x- and y-axes by up to 10\% of the image width or height, and (3) random horizontal and vertical flipping. We evaluated the model's performance on the ImageCLEF 2016 dataset using precision, recall, and F-score metrics.

\begin{table}[t]
\caption{Total number of dermatopathology and non-dermatopathology images in the ImageCLEF2016 dataset used for training and validation of the AI image modality classifier used in this study.\label{tab:clef}}
\centering
\begin{tabular}{lrr}
\toprule
& Training & Test\\
\midrule
Dermatopathology & 696 & 405\\
Other image modalities & 6,080 & 3,761\\
\quad \textit{Total} & 6,776 & 4,166\\
\bottomrule
\end{tabular}
\end{table}

\paragraph{Development of a keyword-based classifier.} 

To ensure the highest quality figures for each diagnosis, we implemented an additional filtering step when more than 10 images were retrieved for a particular diagnosis of interest. In these cases, we further screened pathology images based on the presence of specific keywords, such as ``H and E'', in the figure legends (\textbf{Table \ref{tab:keyword}}). If any of these keywords were found in the corresponding figure legend or manuscript title, the image was deemed relevant to that diagnosis.

\begin{table}
\caption{Keywords used to identify dermatopathology images based on their corresponding figure legends.\label{tab:keyword}}
\centering
\begin{tabular}{ll|ll}
\toprule
H \& E & x4 & Low power & x10 \\
H and E & x4 & Lower power & x10 \\
H\&E & 4x & Lower-power & x50 \\
hematoxylin and eosin & 4x & Nuclei & x50 \\
hematoxylin-eosin & 50x & Pathologic & x20 \\
high magnification & 100x & Pathology & x20 \\
High power & 10x & Pattern & 50x \\
Higher power & 10x & Immunofluorescence & x200 \\
Higher-power & 100x &  & 200x \\
Histopathologic & 20x &  & x200 \\
Histopathology & 20x &  & 400x \\
IHC & 200x &  & x400 \\
Immunohistochemistry & x100 &  & 400x \\
low magnification & x100 &  & x400\\
\bottomrule
\end{tabular}
\end{table}

\paragraph{Development of a hybrid classifier.} 

Given that more images are retrieved for specific diagnoses than others (depending on the rarity of the diagnoses), we further combined the DL classifier with keyword-based classification. A common diagnosis is one that corresponds to equal or more than 10 images retrieved for the diagnosis, and a less common diagnosis is one with fewer than 10 images retrieved. For common diagnoses, we required retrieved images to be identified by both a DL classifier \emph{\textbf{AND}} keyword-based retrieval. For less common diagnoses, we attempted to maximize the number of retrieved images by relaxing the selection process to allow the retrieved image to be identified by either the DL classifier \emph{\textbf{OR}} keywords.

\paragraph{Quality control of a human-labeled dermopathology gold standard.}

To evaluate the performance of the hybrid algorithm, we created a human-labeled gold-standard dataset containing 651 images across six different diagnoses: deep penetrating nevus, fibroepithelioma of Pinkus, glomus tumor, kaposiform hemangioendothelioma, spiradenoma, and warty dyskeratoma. This dataset enabled us to differentiate between pathology images and non-pathology images. 
Two board-certified dermatopathologists validate the gold-standard dataset. As they have a relatively simple task of differentiating between pathology images and non-pathology images, the inter-reader agreement is 100\%.

\paragraph{Performance assessment using the human-labeled dermopathology gold standard.} 

To evaluate the performance of the hybrid algorithm, we created a human-labeled gold-standard dataset containing 651 images across six different diagnoses: deep penetrating nevus, fibroepithelioma of Pinkus, glomus tumor, kaposiform hemangioendothelioma, spiradenoma, and warty dyskeratoma. This dataset enabled us to differentiate between pathology images and non-pathology images. Images identified as pathology by either the DL-based classifier or the keyword-based approach were compared against this gold standard. We measured the precision, recall, and F-score for each diagnosis to assess the effectiveness of the classification algorithms.

\section*{Data Record}\label{data-record}

The DermpathNet dataset is distributed as a compressed archive containing images (\verb|images.zip|), along with three structured metadata files \cite{xu2025-dermpathnet}. All images are consistently named according to the filename field recorded in the metadata. The file \verb|metadata_figures.csv| links each image to its source article and disease annotation, containing information such as the PubMed Central identifier (pmcid), the associated disease label, the filename, and the original figure caption. The file \verb|metadata_ontology.csv| provides standardized disease information using controlled vocabularies, including disease, category, type, UMLS, preferred Name, MESH, and synonyms. The file \verb|metadata_articles.csv| documents the article-level attributes from which the images were derived, including the pmcid and pmid, article title, publishing journal, and license information. Together, these resources allow users to explore DermpathNet at the level of individual images, disease categories, or source articles, while also ensuring transparent documentation of licensing for all included content.

\begin{figure}
\centering
\includegraphics[width=\linewidth]{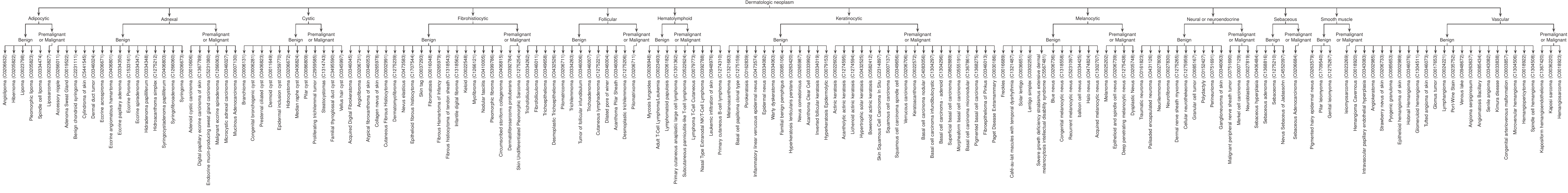}
\caption{Glossary of benign and malignant cutaneous neoplasms of interest for which corresponding articles/images were retrieved and hosted in \DermpathNet.\label{fig:glossary}}
\end{figure}

The DermPath lexicon is an ontology graph consisting of 12 symptom categories, 190 finer-grained Unified Medical Language System (UMLS) concepts, and their synonyms (\textbf{Figure \ref{fig:glossary}}). The list of benign and malignant cutaneous neoplasms of interest includes: keratinocytic, cystic, melanocytic, adnexal, follicular, sebaceous, neural or neuroendocrine, smooth muscle, fibrohistiocytic, adipocytic, vascular, hematolymphoid, and metastatic.

\begin{figure}
\centering
\includegraphics[width=0.9\linewidth]{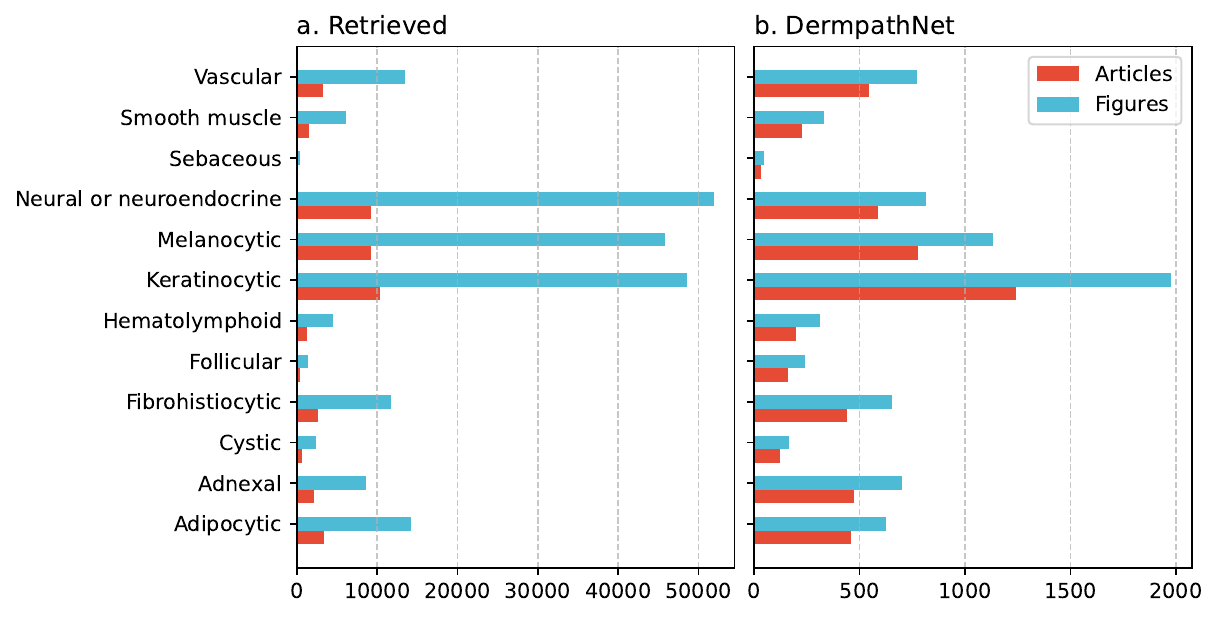}
\caption{Number of articles and images downloaded from PMC-OA pre-filter and published in \DermpathNet amongst different categories of cutaneous neoplasms.\label{fig:articles}}
\end{figure}

We searched the PMC repository to identify articles of interest and determine whether they contained diagnoses of interest in either the title or abstract. We downloaded all corresponding figures, resulting in the curation of 43,821 articles and 209,183 figures \textbf{(Figure \ref{fig:articles}a)}. Most extracted articles and figures fell into the keratinocytic, melanocytic, and neural/neuroendocrine categories, while the least were found in the sebaceous and follicular categories. With the hybrid workflow, we retrieved 7,772 images \textbf{(Figure \ref{fig:articles}b)}. Board-certified dermatopathologists subsequently review all images before being deposited in our dataset.

\section*{Technical validation}

\begin{figure}
\centering
\includegraphics[width=\linewidth]{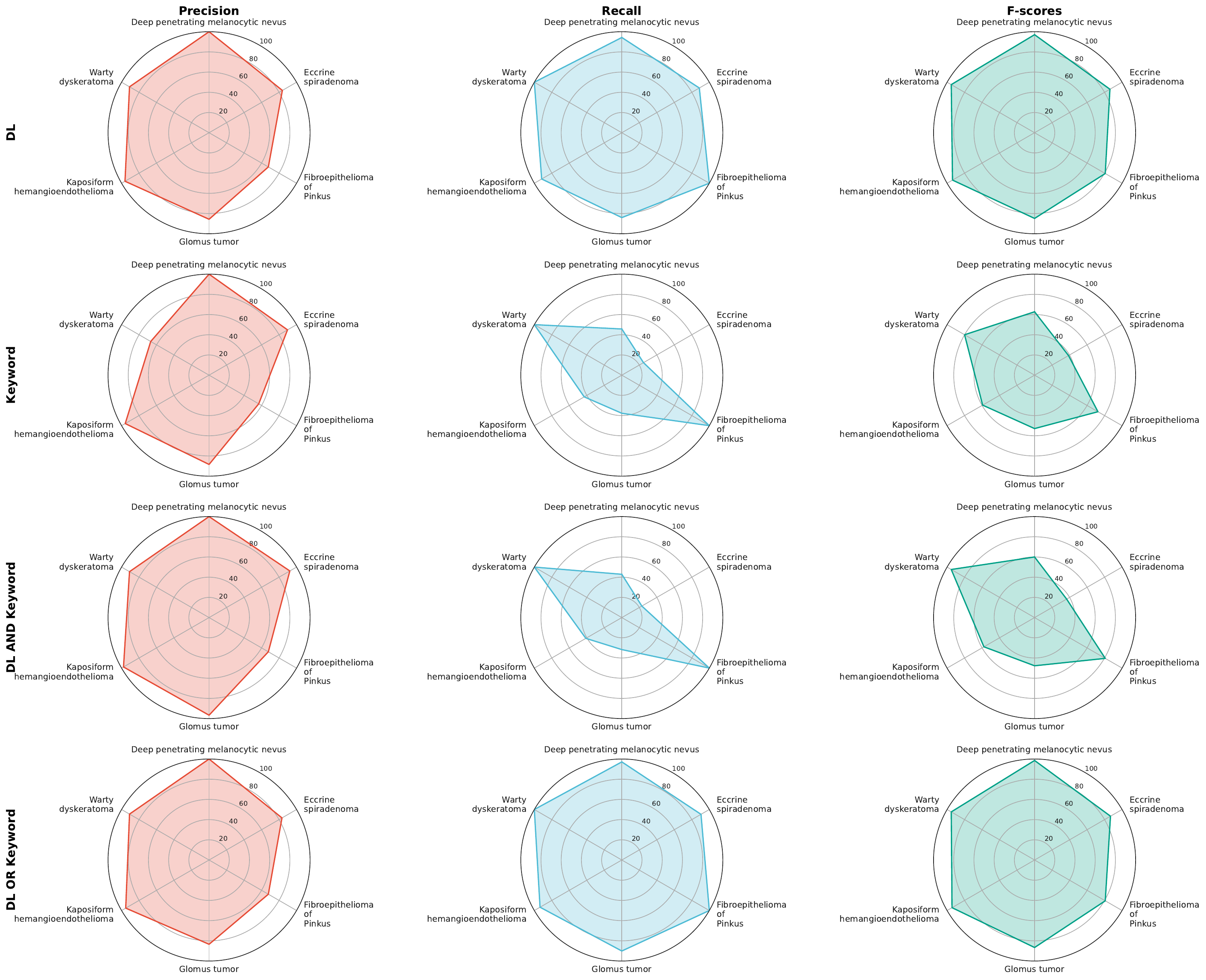}
\caption{Comparison of precision, recall, and F-scores across 6 categories of cutaneous neoplasms.\label{fig:comparison2}}
\end{figure}

We developed two approaches to retrieve pathology images from downloaded PMC image datasets: a DL-based and a keyword-based classifier. \textbf{Figure~\ref{fig:comparison2}} shows that the DL image modality classifier, compared to keyword-based search, retrieved images with similar precision (0.874 vs. 0.823, not significant), higher recall (0.931 vs. 0.585, not significant), and higher F-score (0.896 vs. 0.610, p-value 0.04). Given that more images are retrieved for specific diagnoses than others (depending on the rarity of the diagnoses), we further improved the precision and recall of the retrieval processes by combining the DL classifier with the keyword-based one. For common diagnoses, we required retrieved images to be identified by both a DL classifier \textbf{AND} keywords. This approach results in a precision of 0.909, a recall of 0.564, and an F-score of 0.631. For less common diagnoses, we allowed the retrieved image to be identified by either the DL classifier \textbf{OR} keywords. This alternative approach yields a precision of 0.868, a recall of 0.952, and an F-score of 0.938.

\section*{Discussion}\label{usage-notes}

\subsection*{Evaluating GPT-4v with test images from \DermpathNet.}

\begin{figure}
\centering
\includegraphics[width=0.9\linewidth]{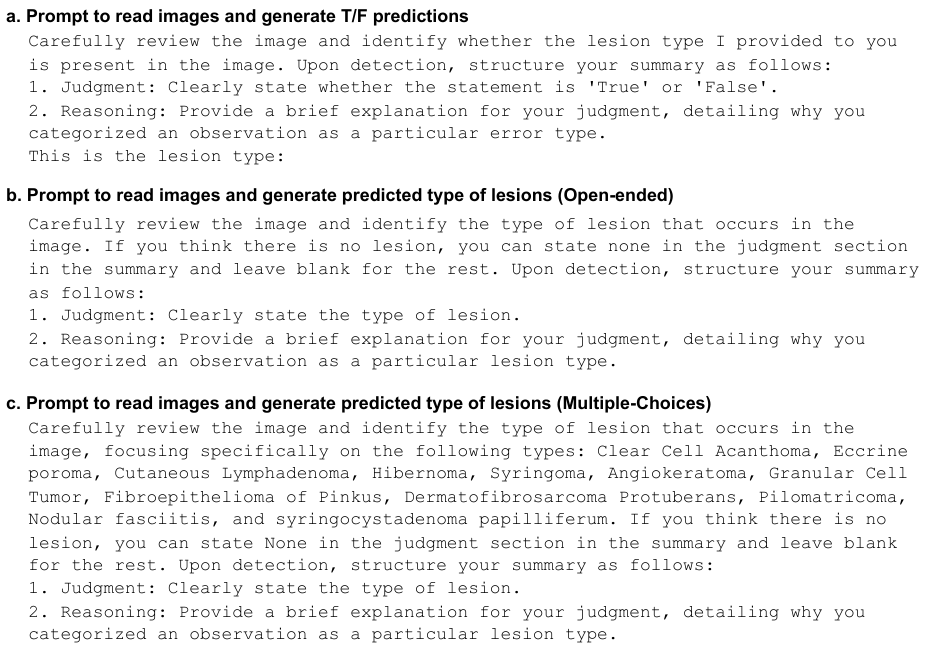}
\caption{Prompts used to assess the image analysis module in GPT-4v in answering True/False, Open-ended, or Multiple-choice questions using randomly selected images from \DermpathNet.\label{fig:prompts}}
\end{figure}

Our dataset was primarily designed for educational use by clinicians, trainees and researchers, enabling exposure to a diverse set of dermatopathology images for learning and cross-referencing. To illustrate the potential of \DermpathNet, we performed exploratory trials using GPT-4v's ability to render correct diagnoses for a set of randomly selected dermatopathology images. Here, we selected 12 different diagnoses of interest, such as angiokeratoma, clear cell acanthoma, and cutaneous lymphadenoma, each representing a distinct type of cutaneous neoplasm. High-resolution images corresponding to these diagnoses were retrieved from \DermpathNet and subsequently validated by board-certified dermatopathologists, resulting in a test dataset of 87 images (\textbf{Table \ref{tab:random}}). We then evaluated GPT-4v's diagnostic accuracy on these images under two different testing scenarios (\textbf{Figure \ref{fig:prompts}}). To prevent GPT-4v from accessing article-identifying information embedded in filenames, all images were renamed prior to evaluation. This ensured that the model’s predictions were based solely on image content rather than metadata such as PubMed identifiers.

\begin{table}
\caption{Number of randomly selected DermpathNet images in each category.\label{tab:random}}
\centering
\begin{tabular}{lr}
\toprule
Category & Number of Images\\
\midrule
Angiokeratoma & 11\\
Granular Cell Tumor & 9\\
Cutaneous Lymphadenoma & 8\\
Dermatofibrosarcoma Protuberans & 8\\
Clear Cell Acanthoma & 7\\
syringocystadenoma Papilliferum & 7\\
Eccrine poroma & 6\\
Hibernoma & 6\\
Syringoma & 6\\
Pilomatrocoma & 6\\
Nodular Fasciitis & 5\\
Fibroepithelioma of Pinkus & 4\\
\quad\textit{Total} & 83\\
\bottomrule
\end{tabular}
\end{table}

\paragraph{Task 1: True/False questions.} 

To construct the dataset, we first split the images of each diagnosis \(d;d \in \{ 1,2,...,6\}\) evenly into two groups: \(\{img_{d}^{1},img_{d}^{2}\}\). For each diagnosis \(d\), the positive samples consisted of \(\{img_{d}^{1}\}\), while the negative samples were randomly selected from \(\bigcup_{d^{'} \neq d}^{}\{ img_{d^{'}}^{2}\}\). This approach ensured an equal balance of positive and negative samples for each diagnostic category. For example, the ``angiokeratoma'' dataset was constructed using 11 images from the angiokeratoma category and 11 additional images randomly chosen from other categories. Each resulting dataset (comprising 22 images in total) was then input into the GPT-4v model (accessed March 30, 2024), along with the prompt: ``Review the image and identify if angiokeratoma occurs in the image.''

\paragraph{Task 2: Open-ended and multiple-choice questions.} 

The same dataset was used to evaluate GPT-4v's performance on open-ended and multiple-choice question formats. Each image was input into GPT-4v with the appropriate prompts (``Review the image and identify the type of lesion that occurs in the image, focusing specifically on {[}list{]}'' and ``Review the image and identify the type of lesion that occurs in the image''). To assess performance, we calculated precision, recall, and F-score for both types of questions. For each image, a true positive was recorded when GPT-4v correctly identified an image with a disease of interest (e.g., angiokeratoma), a false positive when it labeled a non-angiokeratoma image as ``angiokeratoma,'' and a false negative when it failed to identify an angiokeratoma image. These metrics provided a comprehensive assessment of GPT-4v's diagnostic accuracy across both response formats.

\begin{figure}
\centering
\includegraphics[width=0.667\linewidth]{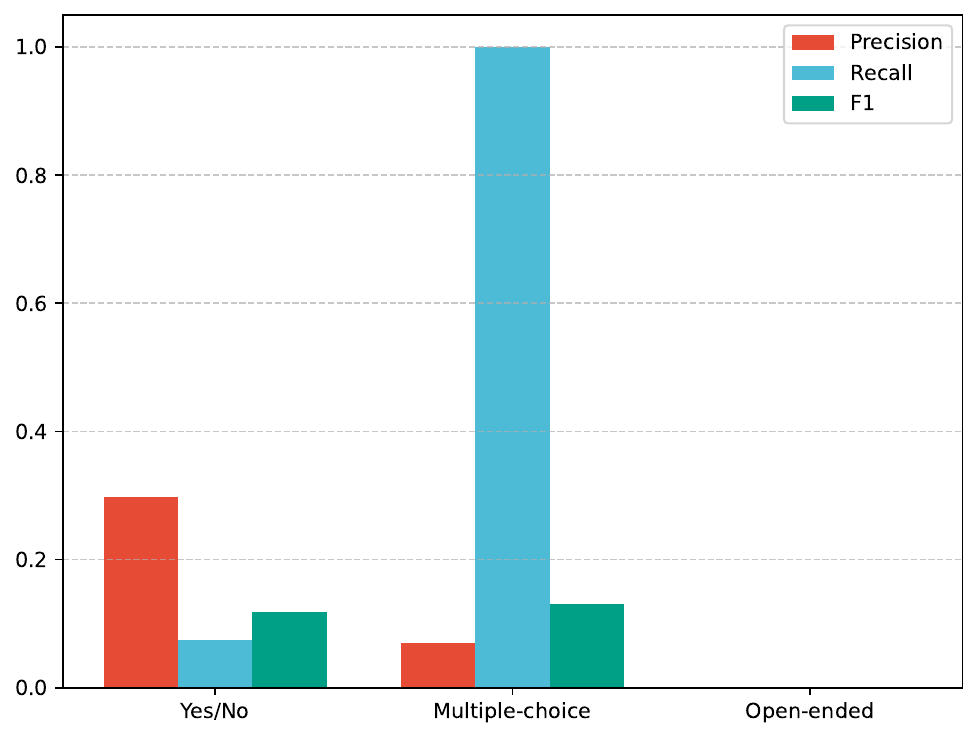}
\caption{Comparison of precision, recall, and F-scores in True/False, Open-ended, and Multiple-choice challenge settings for the image analysis module in GPT-4v using randomly selected dermatopathology images in \DermpathNet.\label{fig:comparison}}
\end{figure}

The GPT-4v model achieved a micro-average precision of 0.30, a recall of 0.07, and an F-score of 0.12 across all 12 categories in the True/false challenge study. Notably, GPT-4v achieved the highest performance in the angiokeratoma category, with an F-score of 0.33. On the other hand, GPT-4v achieved an F-score of 0 when evaluating eccrine poroma, cutaneous lymphadenoma, hibernoma, syringoma, granular cell tumor, fibroepithelioma of pinkus, dermatofibrosarcoma protuberans, or nodular fasciitis. 
In the open-ended challenge, GPT-4v achieved a precision, recall, and F-score of 0 for all diagnoses. In the multiple-choice challenge, it scored a precision of 0.07, a recall of 1, and an F-score of 0.13 (\textbf{Figure \ref{fig:comparison}} and \textbf{Table \ref{stab:comparison}}).

The significant gap in performance metrics was further confirmed when GPT-4v was asked to explain its answers. When the file name contained article-identifying information, GPT-4v relied on clinical information from articles to arrive at the correct diagnosis. After anonymizing file names, GPT-4V continued to show inadequate diagnostic performance, underscoring broader limitations of current multimodal LLMs in this domain. These exploratory findings highlight the pitfalls of relying on contextual artifacts. More positively, they also point to the potential of \DermpathNet to serve not only as an educational resource for clinicians and trainees, but also as a benchmark dataset to guide future development and training of multimodal models for pathology images.

\begin{table}
\caption{Raw data in performance metrics (precision, recall, and F-score) of GPT-4v in True/False questions using randomly selected \DermpathNet images.\label{stab:comparison}}
\centering
\begin{tabular}{lrrr}
\toprule
Disease & Precision & Recall & F-score \\
\midrule
Angiokeratoma & 0.429 & 0.273 & 0.333 \\
Granular Cell Tumor & 0.000 & 0.000 & 0.000 \\
Cutaneous Lymphadenoma & 0.000 & 0.000 & 0.000 \\
Dermatofibrosarcoma Protuberans & 0.000 & 0.000 & 0.000 \\
Clear Cell Acanthoma & 1.000 & 0.143 & 0.250 \\
syringocystadenoma Papilliferum & 1.000 & 0.167 & 0.286 \\
Eccrine Poroma & 0.000 & 0.000 & 0.000 \\
Hibernoma & 0.000 & 0.000 & 0.000 \\
Syringoma & 0.000 & 0.000 & 0.000 \\
Pilomatrocoma & 1.000 & 0.167 & 0.286 \\
Nodular Fasciitis & 0.000 & 0.000 & 0.000 \\
Fibroepithelioma of Pinkus & 0.000 & 0.000 & 0.000 \\
\quad\textit{micro-average} & 0.298 & 0.074 & 0.119\\
\bottomrule
\end{tabular}
\end{table}


\section*{Data Availability}

The data can be found at: \url{https://doi.org/10.5281/zenodo.17288670}.

\section*{Data Provenance and Licensing Statement}

All images in DermpathNet were obtained exclusively from the PubMed Central Open Access Subset (\url{https://pmc.ncbi.nlm.nih.gov/tools/openftlist/}), which includes only articles released under licenses that permit reuse and redistribution.

\section*{Data Maintenance and Versioning Plan Statement}

DermpathNet will be maintained as a living resource with scheduled updates. Each update will incorporate newly available images (if applicable), quality control refinements, and metadata improvements. All changes will be versioned, with detailed changelogs documenting additions or modifications. The dataset will be permanently archived on Zenodo, providing DOI-based access to all versions to ensure long-term accessibility and usability for the research community.

\section*{Code Availability}

The codes can be found at: \url{https://github.com/bionlplab/DermpathNet}.

\section*{Acknowledgements}

This work was supported by the National Science Foundation Career Grant (2145640) and National Institutes of Health (R01LM014306). The funder had no role in the design and conduct of the study; collection, management, analysis, and interpretation of the data; preparation, review, or approval of the manuscript; and decision to submit the manuscript for publication.

\section*{Author Contributions}

Study concepts/study design, \textbf{Z.X., Y.P.}; manuscript drafting or manuscript revision for important intellectual content, all authors; approval of final version of the submitted manuscript, all authors; agrees to ensure any questions related to the work are appropriately resolved, all authors; literature research, \textbf{Z.X.}; experimental studies, \textbf{Z.X., M.L., Y.Z., Z.X.}; human evaluation, \textbf{Z.X., S.J.O., S.A.M.}; data interpretation and statistical analysis, \textbf{Z.X., M.L., Y.Z., Y.P.}; and manuscript editing, all authors.

\section*{Competing Interests}

None declared.

\setlength{\bibsep}{3pt plus 0.3ex}
\bibliographystyle{unsrtnat}
\bibliography{sample}

@ARTICLE{Wen2022-oz,
  title     = "Characteristics of publicly available skin cancer image datasets:
               a systematic review",
  author    = "Wen, David and Khan, Saad M and Ji Xu, Antonio and Ibrahim,
               Hussein and Smith, Luke and Caballero, Jose and Zepeda, Luis and
               de Blas Perez, Carlos and Denniston, Alastair K and Liu, Xiaoxuan
               and Matin, Rubeta N",
  journal   = "Lancet Digit. Health",
  publisher = "Elsevier BV",
  volume    =  4,
  number    =  1,
  pages     = "e64--e74",
  month     =  jan,
  year      =  2022,
  doi       = "10.1016/S2589-7500(21)00252-1",
  pmid      =  34772649,
  issn      = "2589-7500"
}

@ARTICLE{Vardell2012-vl,
  title     = "{VisualDx}: a visual diagnostic decision support tool",
  author    = "Vardell, Emily and Bou-Crick, Carmen",
  journal   = "Med. Ref. Serv. Q.",
  publisher = "Informa UK Limited",
  volume    =  31,
  number    =  4,
  pages     = "414--424",
  year      =  2012,
  doi       = "10.1080/02763869.2012.724287",
  pmid      =  23092418,
  issn      = "0276-3869,1540-9597"
}

@ARTICLE{Beck2011-pz,
  title   = "{NISO} {Z39}.{96The} Journal Article Tag Suite ({JATS}): What
             happened to the {NLM} {DTDs}?",
  author  = "Beck, Jeffrey",
  journal = "J. Electron. Publ.",
  volume  =  14,
  number  =  1,
  year    =  2011,
  doi     = "10.3998/3336451.0014.106",
  pmc     = "PMC3227009",
  pmid    =  22140303,
  issn    = "1080-2711"
}

@ARTICLE{Ward2024-le,
  title   = "Creating an empirical dermatology dataset through crowdsourcing
             with web search advertisements",
  author  = "Ward, Abbi and Li, Jimmy and Wang, Julie and Lakshminarasimhan,
             Sriram and Carrick, Ashley and Campana, Bilson and Hartford, Jay
             and Sreenivasaiah, Pradeep K and Tiyasirisokchai, Tiya and Virmani,
             Sunny and Wong, Renee and Matias, Yossi and Corrado, Greg S and
             Webster, Dale R and Smith, Margaret Ann and Siegel, Dawn and Lin,
             Steven and Ko, Justin and Karthikesalingam, Alan and Semturs,
             Christopher and Rao, Pooja",
  journal = "JAMA Netw. Open",
  volume  =  7,
  number  =  11,
  pages   = "e2446615",
  month   =  "4~" # nov,
  year    =  2024,
  doi     = "10.1001/jamanetworkopen.2024.46615",
  pmc     = "PMC11579799",
  pmid    =  39565619,
  issn    = "2574-3805"
}

@ARTICLE{Schuffler2021-ik,
  title     = "Integrated digital pathology at scale: A solution for clinical
               diagnostics and cancer research at a large academic medical
               center",
  author    = "Sch{\"{u}}ffler, Peter J and Geneslaw, Luke and Yarlagadda, D
               Vijay K and Hanna, Matthew G and Samboy, Jennifer and Stamelos,
               Evangelos and Vanderbilt, Chad and Philip, John and Jean,
               Marc-Henri and Corsale, Lorraine and Manzo, Allyne and
               Paramasivam, Neeraj H G and Ziegler, John S and Gao, Jianjiong
               and Perin, Juan C and Kim, Young Suk and Bhanot, Umeshkumar K and
               Roehrl, Michael H A and Ardon, Orly and Chiang, Sarah and Giri,
               Dilip D and Sigel, Carlie S and Tan, Lee K and Murray, Melissa
               and Virgo, Christina and England, Christine and Yagi, Yukako and
               Sirintrapun, S Joseph and Klimstra, David and Hameed, Meera and
               Reuter, Victor E and Fuchs, Thomas J",
  journal   = "J. Am. Med. Inform. Assoc.",
  publisher = "Oxford University Press (OUP)",
  volume    =  28,
  number    =  9,
  pages     = "1874--1884",
  month     =  "13~" # aug,
  year      =  2021,
  doi       = "10.1093/jamia/ocab085",
  pmc       = "PMC8344580",
  pmid      =  34260720,
  issn      = "1067-5027,1527-974X"
}

@ARTICLE{Moore2015-gl,
  title     = "De-identification of medical images with retention of scientific
               research value",
  author    = "Moore, Stephen M and Maffitt, David R and Smith, Kirk E and
               Kirby, Justin S and Clark, Kenneth W and Freymann, John B and
               Vendt, Bruce A and Tarbox, Lawrence R and Prior, Fred W",
  journal   = "Radiographics",
  publisher = "Radiological Society of North America (RSNA)",
  volume    =  35,
  number    =  3,
  pages     = "727--735",
  month     =  may,
  year      =  2015,
  doi       = "10.1148/rg.2015140244",
  pmc       = "PMC4450976",
  pmid      =  25969931,
  issn      = "0271-5333,1527-1323"
}

@ARTICLE{comeau2019pmc,
  title   = "{PMC} text mining subset in {BioC}: about three million full-text
             articles and growing",
  author  = "Comeau, Donald C and Wei, Chih-Hsuan and Islamaj Do\u{g}an, Rezarta
             and Lu, Zhiyong",
  journal = "Bioinformatics",
  volume  =  35,
  number  =  18,
  pages   = "3533--3535",
  month   =  "15~" # sep,
  year    =  2019,
  doi     = "10.1093/bioinformatics/btz070",
  pmc     = "PMC6748740",
  pmid    =  30715220,
  issn    = "1367-4803,1367-4811"
}

@ARTICLE{Holub2023-hp,
  title   = "Privacy risks of whole-slide image sharing in digital pathology",
  author  = "Holub, Petr and M{\"{u}}ller, Heimo and B\'{\i}l, Tom\'{a}\v{s} and
             Pireddu, Luca and Plass, Markus and Prasser, Fabian and
             Schl{\"{u}}nder, Irene and Zatloukal, Kurt and Nenutil, Rudolf and
             Br\'{a}zdil, Tom\'{a}\v{s}",
  journal = "Nat. Commun.",
  volume  =  14,
  number  =  1,
  pages   =  2577,
  month   =  "4~" # may,
  year    =  2023,
  doi     = "10.1038/s41467-023-37991-y",
  pmc     = "PMC10160114",
  pmid    =  37142591,
  issn    = "2041-1723"
}

@ARTICLE{Feit2005-pg,
  title     = "Hypertext atlas of dermatopathology with expert system for
               epithelial tumors of the skin",
  author    = "Feit, Josef and Kempf, Werner and Jedlickov\'{a}, Hana and Burg,
               G{\"{u}}nter",
  journal   = "J. Cutan. Pathol.",
  publisher = "Wiley",
  volume    =  32,
  number    =  6,
  pages     = "433--437",
  month     =  jul,
  year      =  2005,
  doi       = "10.1111/j.0303-6987.2005.00291.x",
  pmid      =  15953378,
  issn      = "0303-6987,1600-0560"
}

@ARTICLE{Pernick2021-wb,
  title     = "What's new in pathology newsletter by {PathologyOutlines}.Com",
  author    = "Pernick, Nat",
  journal   = "J. Pathol. Transl. Med.",
  publisher = "The Korean Society of Pathologists and The Korean Society for
               Cytopathology",
  volume    =  55,
  number    =  2,
  pages     = "159--160",
  month     =  mar,
  year      =  2021,
  doi       = "10.4132/jptm.2021.03.08",
  pmc       = "PMC7987519",
  pmid      =  33752277,
  issn      = "2383-7837,2383-7845"
}

@ARTICLE{deHerrera2016overview,
  title   = "Overview of the medical tasks in {ImageCLEF} 2016",
  author  = "De Herrera, Alba G Seco and Bromuri, Stefano and Schaer, Roger and
             M{\"{u}}ller, Henning",
  journal = "CLEF Working Notes. Evora, Portugal",
  year    =  2016
}

@ARTICLE{Lee2015-kq,
  title   = "{iSlide}: a 'big picture' interactive teledermatopathology
             e-learning system",
  author  = "Lee, P and Chen, C-F and Wan, H-T and Jian, W-S and Hsu, M-H and
             Syed-Abdul, S and Huang, C-W and Huang, Y-C and Lin, Y-T and Chen,
             T-J and Wu, Y-H and Li, Y-C",
  journal = "Br. J. Dermatol.",
  volume  =  172,
  number  =  3,
  pages   = "692--699",
  month   =  mar,
  year    =  2015,
  doi     = "10.1111/bjd.13274",
  pmid    =  25040884,
  issn    = "0007-0963,1365-2133"
}

@INCOLLECTION{Yao2021-ik,
  title     = "Compound figure separation of biomedical images with side loss",
  author    = "Yao, Tianyuan and Qu, Chang and Liu, Quan and Deng, Ruining and
               Tian, Yuanhan and Xu, Jiachen and Jha, Aadarsh and Bao, Shunxing
               and Zhao, Mengyang and Fogo, Agnes B and Landman, Bennett A and
               Chang, Catie and Yang, Haichun and Huo, Yuankai",
  booktitle = "Lecture Notes in Computer Science",
  publisher = "Springer International Publishing",
  address   = "Cham",
  pages     = "173--183",
  series    = "Lecture notes in computer science",
  year      =  2021,
  doi       = "10.1007/978-3-030-88210-5\_16",
  isbn      = "9783030882099,9783030882105",
  issn      = "0302-9743,1611-3349"
}

@ARTICLE{Ossom_Williamson2019-hy,
  title     = "Exploring {PubMed} as a reliable resource for scholarly
               communications services",
  author    = "Ossom Williamson, Peace and Minter, Christian I J",
  journal   = "J. Med. Libr. Assoc.",
  publisher = "University Library System, University of Pittsburgh",
  volume    =  107,
  number    =  1,
  pages     = "16--29",
  month     =  jan,
  year      =  2019,
  doi       = "10.5195/jmla.2019.433",
  pmc       = "PMC6300231",
  pmid      =  30598645,
  issn      = "1536-5050,1558-9439"
}

@INPROCEEDINGS{deng2009imagenet,
  title     = "{ImageNet}: A large-scale hierarchical image database",
  author    = "Deng, Jia and Dong, Wei and Socher, Richard and Li, Li-Jia and
               Li, Kai and Fei-Fei, Li",
  booktitle = "2009 IEEE Conference on Computer Vision and Pattern Recognition",
  pages     = "248--255",
  month     =  jun,
  year      =  2009,
  doi       = "10.1109/CVPR.2009.5206848",
  issn      = "1063-6919"
}

@ARTICLE{Rinck2023-mb,
  title     = "National resident survey in dermatopathology: The role of slide
               scanners in resident learning",
  author    = "Rinck, Danielle and Dittmer, Martin and Tinker, Daniel and Smith,
               Kristin and Heinecke, Gillian",
  journal   = "J. Cutan. Pathol.",
  publisher = "Wiley",
  volume    =  50,
  number    =  12,
  pages     = "1078--1082",
  month     =  dec,
  year      =  2023,
  doi       = "10.1111/cup.14538",
  pmc       = "PMC10843035",
  pmid      =  37749824,
  issn      = "0303-6987,1600-0560"
}

@MISC{sayers2022general,
  title        = "A General Introduction to the {E}-utilities",
  author       = "Sayers, Eric",
  booktitle    = "Entrez Programming Utilities Help [Internet]",
  publisher    = "National Center for Biotechnology Information (US)",
  month        =  "17~" # nov,
  year         =  2022,
  howpublished = "\url{https://www.ncbi.nlm.nih.gov/books/NBK25497/}"
}

@ARTICLE{Brown2015-zd,
  title     = "Uniform labeling of blocks and slides in surgical pathology:
               Guideline from the College of American Pathologists Pathology and
               Laboratory Quality Center and the National Society for
               Histotechnology",
  author    = "Brown, Richard W and Della Speranza, Vincent and Alvarez, Janice
               O and Eisen, Richard N and Frishberg, David P and Rosai, Juan and
               Santiago, Jerry and Tunnicliffe, Janet and Colasacco, Carol and
               Lacchetti, Christina and Thomas, Nicole E",
  journal   = "Arch. Pathol. Lab. Med.",
  publisher = "Archives of Pathology and Laboratory Medicine",
  volume    =  139,
  number    =  12,
  pages     = "1515--1524",
  month     =  dec,
  year      =  2015,
  doi       = "10.5858/arpa.2014-0340-SA",
  pmid      =  25897820,
  issn      = "0003-9985,1543-2165"
}

@ARTICLE{Shahriari2017-pg,
  title   = "Dermatopathology education in the era of modern technology",
  author  = "Shahriari, Neda and Grant-Kels, Jane and Murphy, Michael J",
  journal = "J. Cutan. Pathol.",
  volume  =  44,
  number  =  9,
  pages   = "763--771",
  month   =  sep,
  year    =  2017,
  doi     = "10.1111/cup.12980",
  pmid    =  28612388,
  issn    = "0303-6987,1600-0560"
}

@ARTICLE{xu2025-dermpathnet,
    title = "{DermpathNet}",
    author = "Ziyang Xu and Mingquan Lin and Yiliang Zhou and Zihan Xu and Seth J. Orlow and Shane A. Meehan and Alexandra Flamm and Ata S. Moshiri and Yifan Peng",
    journal = "Zenodo",
    doi = "10.5281/zenodo.17288670",
    year = 2025,
    publisher = "Zenodo",
}

\end{document}